\documentclass[letterpaper, 10 pt, conference]{ieeeconf}  
\usepackage[T1]{fontenc}

\IEEEoverridecommandlockouts                              

\usepackage{graphicx} 
\usepackage{xcolor}
\usepackage{url}
\usepackage{booktabs}
\usepackage{listings}
\usepackage{hyphenat}
\usepackage[english]{babel}

\title{\LARGE \bf
Would you let a humanoid play storytelling with your child? A usability study on LLM-powered narrative Human-Robot Interaction}

\author{Maria Lombardi$^{1,\ast,\ddagger}$, Carmela Calabrese$^{1,\ast,\ddagger}$, Davide Ghiglino$^{2}$, Caterina Foglino$^{2}$, \\ Davide De Tommaso$^{2}$, Giulia Da Lisca$^{2}$, Lorenzo Natale$^{1, \S}$, Agnieszka Wykowska$^{2, \S}$
\thanks{$\ast$Shared first authorship, $^{\S}$Shared last authorship}
\thanks{$\ddagger$Corresponding authors: {\tt\small maria.lombardi1@iit.it, carmela.calabrese@iit.it}}
\thanks{$^{1}$M.L., C.C., and L.N. are with Humanoid Sensing and Perception, Italian Institute of Technology (IIT), Genoa, 16163, Italy}%
\thanks{$^{2}$D.G., C.F., D.D.T., G.D.L., and A.W. are with Social Cognition in Human-Robot Interaction, IIT, Genoa, 16163, Italy}%
\thanks{This work received funding from the project Fit for Medical Robotics (Fit4MedRob) - PNRR MUR Cod. PNC0000007 - CUP: B53C22006960001 and from the European Research Council (ERC) under the European Union’s Horizon Europe research and innovation programme (Grant title ``RONIN - Robot Training Independence'', agreement No. ERC-2023-POC-101155938, awarded to AW.}
}

\begin{document}

\maketitle
\thispagestyle{empty}
\pagestyle{empty}

\begin{abstract}
A key challenge in human-robot interaction research lies in developing robotic systems that can effectively perceive and interpret social cues, facilitating natural and adaptive interactions. In this work, we present a novel framework for enhancing the attention of the iCub humanoid robot by integrating advanced perceptual abilities to recognise social cues, understand surroundings through generative models, such as ChatGPT, and respond with contextually appropriate social behaviour. Specifically, we propose an interaction task implementing a narrative protocol (storytelling task) in which the human and the robot create a short imaginary story together, exchanging in turn cubes with creative images placed on them. To validate the protocol and the framework, experiments were performed to quantify the degree of usability and the quality of experience perceived by participants interacting with the system. Such a system can be beneficial in promoting effective human–robot collaborations, especially in assistance, education and rehabilitation scenarios where the social awareness and the robot responsiveness play a pivotal role.
\end{abstract}

\section{INTRODUCTION}
Humanoid robots with social capabilities have significant potential in diverse fields, including elderly care and healthcare systems. A key challenge in Human-Robot Interaction (HRI) research is developing a robust system that can perceive and recognize various social cues, enabling natural and effective interactions between humans and robots \cite{holman2021watch}. Visual perception allows robots to understand their surroundings, anticipate human actions, and react appropriately during collaborative tasks. On the other hand, generative models like ChatGPT \cite{chatgpt} and vision-language models (VLMs) \cite{liu2024visual} significantly improve the fluidity and naturalness of these interactions by enabling contextually appropriate and adaptive communication in real-time, and so facilitating a less rigid interaction. In this context, we exploit the capabilities of the generative models to propose a human-robot interaction task implementing a storytelling protocol in which human and robot have to create a story together in turn, inspired by a set of pictures placed on soft cubes.

Storytelling is one of the oldest forms of self-expression, a way that people can relate to each other and give meaning to everyday experiences. Recent studies highlight the potential of integrating Large Language Models (LLMs) and Socially Assistive Robots (SAR) in narrative protocols to improve the efficacy of cognitive rehabilitation and therapeutic education by providing personalized support to individuals. For example, narrative tasks may help children with social disorders to explore their feelings and develop language to express their thoughts \cite{berrezueta2024exploring}. Therefore, building on the importance of attention and adaptability in social cognition, we equip the iCub humanoid robot \cite{metta2010icub} with advanced perceptual abilities to recognise and respond to social cues from human partners. 


This framework improves iCub’s social awareness and responsiveness, its ability to adjust responses to individual user's needs and conversational flows, leading to more engaging and human-like exchanges, particularly crucial in educations and in rehabilitation applications \cite{alabdulkareem2022systematic, calderita2014therapist}. In this context, we propose a storytelling system mediated by imagery-based cubes, for applications in protocols suitable for both educational and rehabilitation purposes, for example training social skills in children diagnosed with autism spectrum disorder (ASD). In this study with adult subjects, we investigate the system in terms of usability, acceptability and perceived usefulness as first and necessary step before deploying the robotic system in interaction with children.

\section{STATE OF THE ART}
\subsection{Humanoids in social and cognitive rehabilitation}
The intersection of robotics and healthcare is rapidly evolving, with humanoid robots emerging as promising tools for various rehabilitation scenarios, from physical (\cite{calabrese2025socially,feingold2021robot}) to cognitive rehabilitation (\cite{zheng2020randomized,morris2024enhancing}). By providing consistent, predictable, and engaging interactions, humanoids can address a range of challenges faced by individuals recovering from cognitive impairments, neurological conditions, or social difficulties. One of the most popular robots used especially in social rehabilitation today is NAO.
For example in \cite{david2018developing}, authors investigated the events of joint attention between the ASD children and the robot manipulating 3 different social cues: gaze orientation, pointing and vocal instruction. They found that the most engaging social cue was the pointing. The robot's movement was controlled by the operator through a Wizard of Oz paradigm (WoZ). 

Another widely used robot is Pepper, which is equipped with facial recognition capabilities to read emotions and engage in conversation. It can perform various tasks and provide information through direct interaction via a touchscreen. While NAO is mainly used in ASD rehabilitation, several studies can be found using Pepper to assist individuals with schizophrenia or dementia. For example, in \cite{morris2024enhancing} Pepper was used to enhance dementia care through music therapy. The Pepper's tablet was used to display buttons, lyrics and other visual stimuli during the task. The robot itself serves as supporter giving positive feedback and as dance companion by using pre-programmed upper body animation. 


Across many studies, humanoid robots, despite their varying forms, features, and abilities, have been used mainly for monitoring or as facilitators during training, providing negative/positive feedback. Furthermore, all the aforementioned studies used either a robot programmed a-priori or teleoperated by using WoZ paradigm or, if any, a perception system that is external to the robot \cite{zheng2020randomized}.

In our study, we used the humanoid robot iCub \cite{metta2010icub} which balances human-likeness while avoiding the uncanny valley effect \cite{mori2012uncanny}. In addition, unlike other types of humanoids, previous studies shown that its human-likeness with its capacity to mimic human behaviour have potential to engage children during the interaction, making it an ideal choice for our research \cite{ghiglino2023artificial}.
We equipped iCub with a perception system able to sense the surrounding environment to fulfill the given storytelling task, taking into account the feedback received from the surroundings. Furthermore, to increase the ecological validity of the interaction, all the implemented perception algorithms exploit the RGB frames coming from the iCub's eye without using any external sensor.

\subsection{Foundation models for AI-powered storytelling systems}

Narrative architectures are often used to foster empathy and connection towards other people. The personality of the robot, such as being encouraging or non-judging, can help facilitating emotional engagement in everyday life \cite{shen2025social}. LLMs can serve as a key component in the design of social robots, enhancing HRI.
Embodied social robots equipped with complex conversational capabilities are perceived differently in emotion attribution and understanding \cite{de2024you}. Adopting more advanced chatting capabilities based on LLMs significantly enhances the perception of robot's emotional expression, as well as its clarity and expressiveness, with structured dialogue further boosting engagement \cite{kang2024nadine}.

Pretending storytelling with parents is crucial for child development, supporting language, social, and emotional growth. Studies comparing LLMs, such as ChatGPT, with parental interactions to engage children in creative storytelling suggest that children largely enjoy and even prefer interacting with ChatGPT, drawn by the novelty of the experience and its collaborative narrative approach \cite{chin2024young}.  
Thus, narrative architecture can aid in creating therapy protocols to improve communication skills, turn taking, and autonomy, particularly in autistic children. Storytelling based on specific scenarios and cues can encourage them to complete tasks and create their own narratives \cite{chatzara2014digital}, helping them navigate challenges related to activity transitions. With their adaptability, LLMs and AI-based solutions provide promising customized solutions through real-time interactions, reducing the burden on educators \cite{cao2025designing,lyu2024designing}.

In human-child story co-creation, children often create stories while playing with objects or toys in their hands \cite{talu2018symbolic}, allowing them to express ideas through gestures rather than just language, particularly useful for impaired players struggling to verbalize. Therefore, enabling LLMs to process both speech and visual inputs - such as children’s drawings, gestures, and facial expressions - can lead to more natural and immersive interactions. Emerging AI system, like Toyteller \cite{chung2025toyteller}, support collaborative multimodal storytelling, allowing LLMs to process character symbol movements as both input and output, enabling users to co-create stories by guiding character interactions and movements.
For these reasons, foundation models like LLMs, VLMs, or generative models for text-to-image applications, play today a key-role in  making the interaction more engaging. 
Additionally, advanced embodied AI agents that integrate LLMs with emotion and intent recognition modules provide more context-aware responses, moving beyond simple reactions. This allows for dynamic personality adaptations, enhancing collaborative storytelling and fostering creativity, engagement, and user satisfaction \cite{jia2025embodied}. 


\section{DESIGN}\label{sec:design}
\begin{figure}[h]
   \centering
          \includegraphics[width=0.5\columnwidth]{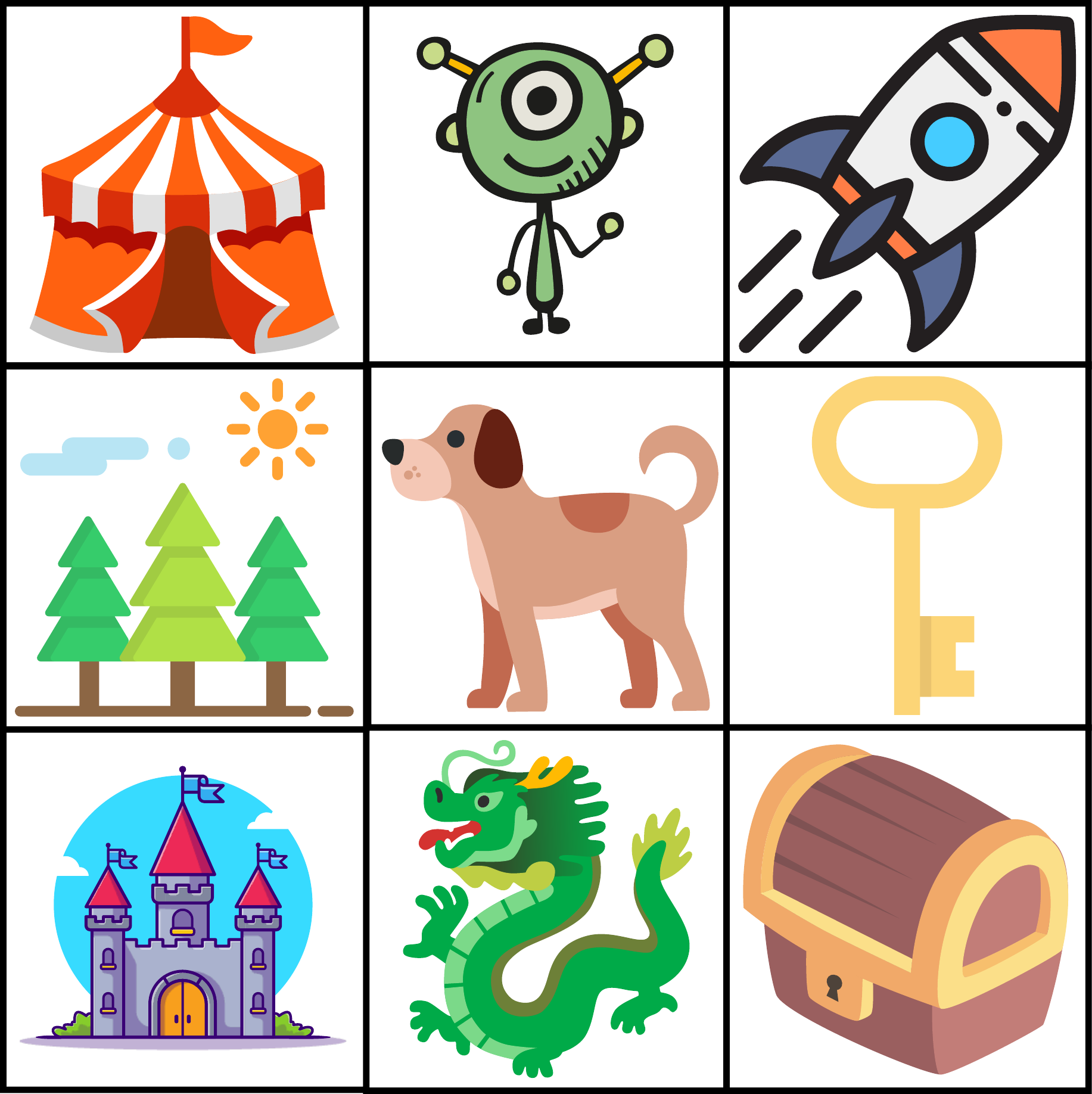}
   \caption{\textbf{Stickers.} A total of 9 stickers were used for the storytelling, combining three places, three characters and three objects.}
   \label{fig:stickers}
\end{figure}

We designed a companion-like social robotic platform able to support users in narrative tasks.
During the interaction, the participant and the iCub robot face each other and have to create stories together, in turn, starting from a set of given pictures. The visual stimuli are placed on nine soft cubes with stickers representing different objects/characters (e.g. castle, alien, see Fig. \ref{fig:stickers}). Each story is built on the sequence of three cubes. Furthermore, the participant is equipped with a microphone to acquire their part of the story. 

\begin{figure}[h]
   \centering
          \includegraphics[width=0.7\columnwidth]{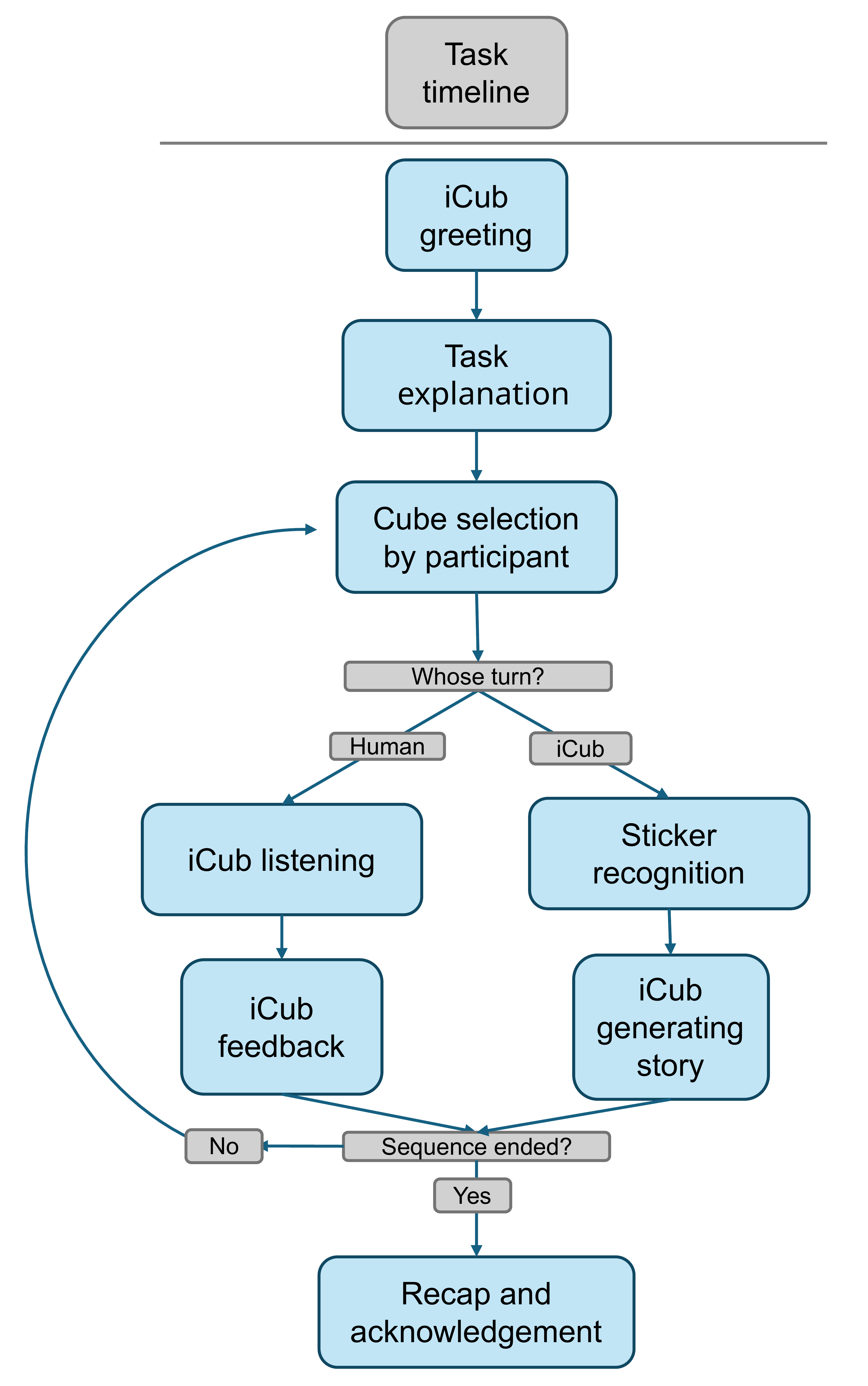}
   \caption{\textbf{Interaction timeline.} The flow diagram describes the different phases of the storytelling task.}
   \label{fig:timeline}
\end{figure}

We structured the human-robot interaction in the following phases (see Fig. \ref{fig:timeline}):
\begin{enumerate}
    \item \textbf{Introduction phase.} At the beginning of each section, the robot describes the context and gives details on the task. In this phase, the robot detects the participant with whom it is interacting. 
    \item \textbf{iCub Turn (Beginning of the trial).} The robot asks the participant to choose and pass it the first cube. In this phase, the robot grasps, detects, and identifies the sticker on the cube, and starts the story based on that image.
    \item \textbf{Human Turn.} The participant chooses the second cube and continues the story while listened by iCub. Once invented their part of the story, the participant passes the second cube to iCub allowing it to see the cube and give a (positive) feedback.
    \item \textbf{iCub Turn.} iCub asks again the participant to choose and pass it the final cube to create an ending to the story.
    \item \textbf{Wrap-up Phase (Ending of the trial).} In this step, iCub shares a recap of the jointly generated story to the user. 
    \item \textbf{Closure Phase.} iCub greets the participant, expressing joy.
\end{enumerate}
Phases 2-5 can be repeated as many times as the number of experimental trials.
\vspace{-1mm}
\vspace{-3mm}

\section{SYSTEM ARCHITECTURE}
\subsection{iCub robotic platform}
The humanoid robot iCub was used as main robotic platform to implement and validate the designed HRI protocol~\cite{Metta2010}. The iCub robot was connected over a local network to a workstation acting as a server and another client laptop, both equipped with external GPUs (NVIDIA GeForce GTX 1080). Specifically, the workstation was used to control and launch the experiments over the clients, to interact and monitor the experiment through the user interface, to control the robot's behaviour and run the modules related to speech and text generation. The other client, instead, was used to run all the modules related to the robot's perception (e.g. object detection, face recognition, mutual gaze). All the running modules were connected through the robotic middleware YARP~\footnote{\url{https://www.yarp.it}}, an open-source framework providing protocols for peer-to-peer communication between different applications.

\subsection{iCub robot perception and interaction system}
\begin{figure*}[t]
    \centering
    \includegraphics[width=0.9\textwidth]{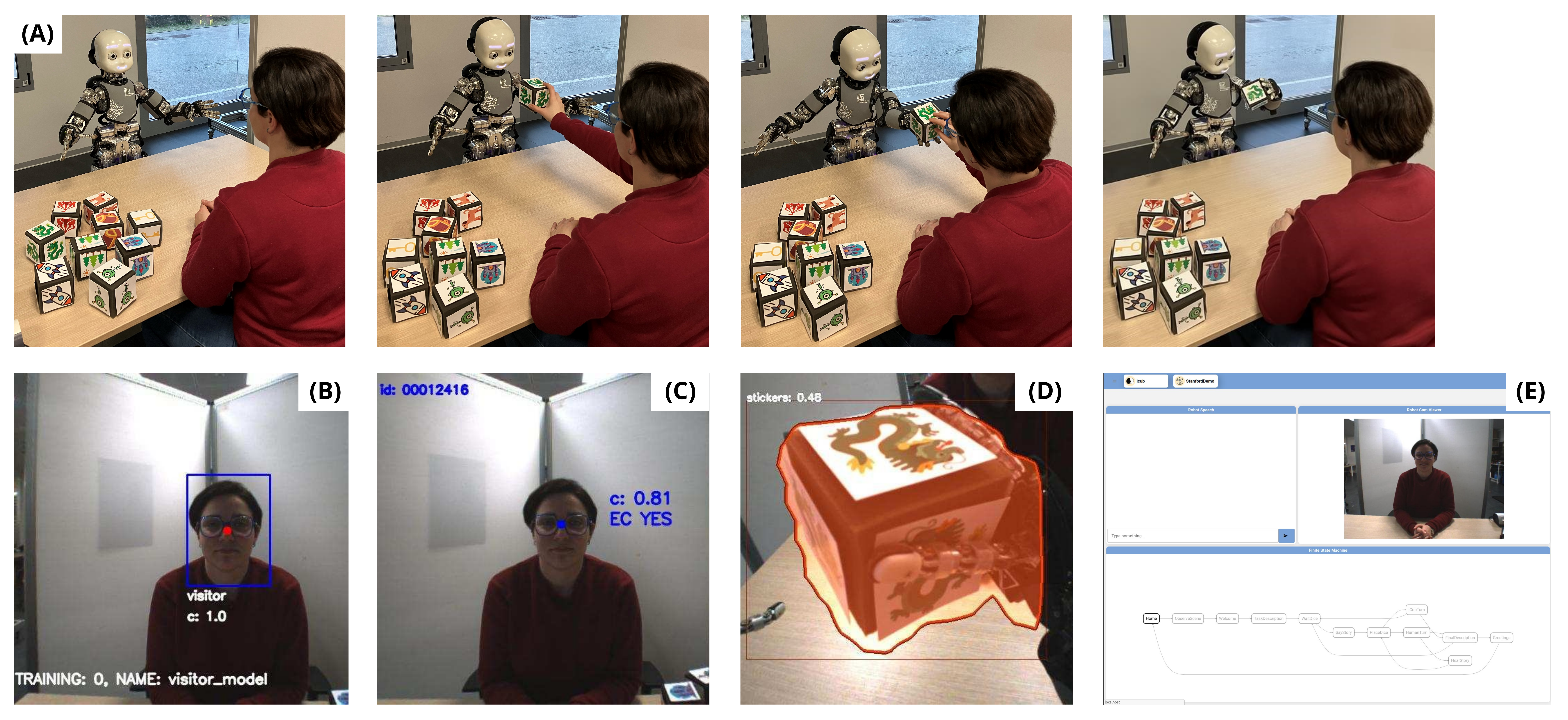}
    \caption{\textbf{System Components and Interaction Flow.} (A) Sequence of human-iCub interaction during the cube handover.  (B) Face recognition and tracking. (C) Mutual gaze. (D) Soft cube detected by the object detector. (E) Overview of the dedicated user interface it was used by the experimenter for real time monitoring.}
   \label{fig:modules}
\end{figure*}

Different perception and manipulation modules were employed to program the iCub robot to fulfill the task with the human partner correctly (see Fig.\ref{fig:modules}). All the modules are described in what follows.

\vspace{-5mm}

\subsubsection{Face recognition}
The face recognition module aims at recognising the human the robot has to interact with during the task to allow it to track and look at the specific human partner in case of multiple people in the scene (Fig.\ref{fig:modules}-(\textbf{B})). To this aim, we used a revised version of the module proposed in~\cite{Lombardi2022a}. Briefly, once the module receives the RGB frame from iCub's camera embedded in its right eye, OpenPose~\cite{Cao2019} is used as pose estimator to build the bounding box around the face of each individual in the image. 
The thumbnails are given first to the pre-trained network FaceNet~\cite{Schroff2015} to extract the face embeddings and then to a binary Support Vector Machine (SVM) classifier to recognise the human partner of interest from the other people. 
In case the face recognition module fails in the prediction, the closest bounding box is returned. This is done by the assumption that the human interacting with the robot is the closest to it. 
This module works completely online with the learning phase running at the beginning of the interaction, automatically annotating the biggest bounding box as ground truth.
The main advantage to have a face recognition component that can be trained online is that the system is not constrained to a specific subset of pre-trained human partners, making it efficiently work also in real and dynamic scenarios.

\subsubsection{Mutual gaze}
Once the human partner is recognized, we rely on the module proposed in~\cite{Lombardi2022b} to detect events of mutual gaze between they and the iCub robot (Fig.\ref{fig:modules}-(\textbf{C})). Briefly, such a classifier exploits OpenPose to extract the feature vector of the individual of interest (in the proposed application this is identified by the \textit{Face recognition} module). A subset of $19$ facial key-points are considered  ($8$ points for each eye, $2$ points for the ears and $1$ for the nose) resulting in a feature vector of $57$ elements (i.e. the triplet $(x,y,k)$ is taken for each point). Then, a binary SVM classifier is trained to recognise events of \textit{eye-contact / no eye-contact}.

\subsubsection{Object detection}
The soft cubes used during the interaction are detected and recognised using the object detection module proposed in~\cite{Ceola2020} (see Fig.\ref{fig:modules}-\textbf{(D)}). Specifically, given an input image acquired by the iCub's eye, it mainly consists in a two layers architecture: a first layer for the prediction of region proposals and feature extraction, and a second layer for the region classification and bounding box refinement. The former used layers from Mask R-CNN to extract a set of Regions of Interest (RoIs) from the input image and encode them into convolutional features. The latter, instead, classifies and refines the proposed RoIs using a set of binary classifiers (one for each object class to detect) for the classification and Regularized Least Squares (RLS) for the refinement. The main advantage of this object detector is that allow us to train new objects online, allowing for a fast adaptation to a dynamic environment without performance loss (details can be found~\cite{Ceola2020}).

\subsubsection{Sticker detection}
Each cube represented a different visual stimulus to start or give context to the story generation. In this work, we leveraged the zero-shot capabilities of Visual Language Models (VLMs) to retrieve a description of the cartoon. This approach allows to change stimuli fast without training the system again, representing a future easily adaptive tool for therapists. In this case, we used ChatGPT-4o as VLM with an appropriate ad-hoc prompt.

\begin{lstlisting}
"system prompt": "You will cooperate with a narrative LLM to create a story. You will be provided with a small cardboard box with a sticker on it. The sticker can depict various scenarios/animals/object/characters. You are a describer that will be asked to recognize what is inside the sticker with a cartoon. Do not invent, be conservative. You have to be very confident in your answer. Do not be too much wordy, provide short descriptions. As much brief as possible, You must use maximum 10 words. What is inside the sticker? Tell only which character/object you see in the sticker, using only 2 adjectiives, without using the word 'sticker'. E.g. A grey smiling koala or a mushroom house with red roof." 
\end{lstlisting}

\subsubsection{Story generation}
Storytelling is the primary focus of this work. The narrative technique is often a key aspect in the design of systems for autistic children to encourage them to complete tasks and create their own narratives. Certain story scenarios and cues can help improve the communication skills and symbolic functions of autistic children, thereby fostering their autonomy \cite{chatzara2014digital}. The generative capabilities of LLMs can satisfy these requirements, improving variability, and, therefore, fostering engagement. We used ChatGPT-4o as LLM with an appropriate ad-hoc prompt.

\begin{lstlisting}
"system prompt": "You are a humanoid robot called iCub, developped at the Italian Institute of Technology. iCub can emulate many of a 6-8 years-old human capacities. Manipulation, vision, and hearing are its main capacities. You will be asked to invent a story for a 6-8 years-old child. To do this, you will be asked to invent short pieces of a simple complete story in three steps, starting from the description of a scenario. Avoid using words like cartoon, cardbox or sticker. Please, remember to be short- maximum 15 words- and simple, and to create a homogeneous story that ends in 3 steps."
\end{lstlisting}

\subsubsection{Speech transcription and synthesizer}
Storytelling requires iCub to synthesize the story generated by the LLM into verbal signals, and detect and understand the story snippet told by the human, ensuring coherence with the ongoing narrative in the next round. To this end, the iCub robot embeds two extra modules: speech-to-text and text-to-speech. While we use the speech module of YARP framework as speech synthesizer, we leverage speech-to-text technique from the Google Speech API as speech transcription\footnote{\url{https://cloud.google.com/speech-to-text}}.

\subsubsection{Finite State Machine (FSM)}
All previously mentioned perception modules, as well as those responsible for controlling the robot’s sensors, actuators, and controllers, are interconnected through YARP. Their collective behavior is orchestrated by a supervisor process that employs an FSM, with conditional transitions implemented using the \textit{transitions} Python module\footnote{\url{https://github.com/pytransitions/transitions}}. To monitor the process in real time and to debug the evolution of the FSM, a dedicated user interface was used by the experimenter (see Fig.\ref{fig:modules}-\textbf{(E})).

\section{EXPERIMENTS}
\subsection{Participants}
Twenty-six participants were recruited to participate in the validation experiment (mean age = $39.7\pm15$, $16$ females). One participant was excluded from the analyses because of previous participation to multiple experiments with iCub. All other participants were n\"aive to robotics and gave written informed consent to participate in the study. Data collection was carried out in compliance with the Ethical standards of the 2013 Declaration of Helsinki, and the procedures received approval from the regional ethics committee (Comitato Etico Regione Liguria). All participants received an honorarium of $10$ euros for participating.

\subsection{Procedure}
Each experimental session was divided in two phases: \textit{Interaction} and \textit{Post-Interaction}.  During the \textit{Interaction} phase (described in Section \ref{sec:design}), each participant invented three stories (trials) with iCub. At the end of each trial, they were asked to complete a short ad-hoc questionnaire focused on evaluating each jointly generated outcome. During the  \textit{Post-Interaction} phase, participants filled out two usability questionnaires and completed another ad-hoc questionnaire. The overall experiment lasted approximately one hour. 

\subsection{Questionnaires}
We used a set of questionnaires to collect qualitative feedback on the narrative robot iCub's functionalities from the users' perspective in terms of usability, acceptability, and perceived usefulness. In particular, the following questionnaires were used during the two experimental phases:
\subsubsection{During interaction}
\begin{itemize}
    \item \textbf{Ad-hoc questionnaire 1.} To collect feedback on the quality of the stories created, we asked participants to rate the likability and coherence of each story at the end of each trial. The questionnaire consisted of two simple items rated on a $1$–$10$ scale. The items included in the questionnaire are presented in Table \ref{adhoc1}.
\end{itemize}

\subsubsection{Post-interaction}
\begin{itemize}
\item \textbf{System Usability Scale (SUS)} \cite{brooke1996sus}. This questionnaire is highly robust and versatile tool commonly used to assess the global acceptability and usability of technological solutions. It includes $10$ statements and subjects indicate their level of agreement with each item on a $1$–$5$ Likert scale (ranging from \textit{``strongly disagree''} $1$ and \textit{``strongly agree''} $5$). Odd-numbered items are positively worded (e.g., \textit{``I think that I would like to use this system frequently''} and \textit{``I thought the system was easy to use''}), meaning higher ratings indicate better usability. Even-numbered items are negatively worded (e.g., \textit{``I found the system unnecessarily complex''} and \textit{``I think that I would need the support of a technical person to be able to use this system''}), meaning higher ratings indicate worse usability. The overall SUS score is calculated by summing the adjusted item scores and multiplying by $2.5$, resulting in a range of $0$ to $100$. 
\item \textbf{User Experience Questionnaire (UEQ)} \cite{laugwitz2006konstruktion}. The UEQ questionnaire has been used in several studies examining the quality of interactive products. It serves as an analytical approach to ensure the practical relevance of distinct quality aspects, which are summarized in six scales comprising $26$ items: Attractiveness (e.g.,\textit{``annoying / enjoyable''}), Perspicuity (e.g.,\textit{``not understandable / understandable''}), Efficiency (e.g.,\textit{``fast / slow''}), Dependability (e.g.,\textit{``unpredictable / predictable''}), Stimulation (e.g.,\textit{``boring / exciting''}), Novelty (e.g.,\textit{``creative / dull''}). Each item is rated on a $7$-point Likert scale, with resulting responses ranging from $-3$ (strongly agree with the negative statement) to $+3$ (strongly agree with the positive statement). However, in practice, most responses fall within a narrower range (typically between $-2$ and $+2$) \cite{schrepp2017construction}.
\item \textbf{Ad-hoc questionnaire 2}. The second ad-hoc questionnaire consisted in $7$ questions aiming to evaluate participants' experiences and perceptions of their interaction with the robot iCub. This questionnaire enabled us to explore specific aspects of the proposed solution, gather preferences from end-users, and derive valuable insights for the robot's future development. The items of the questionnaire are reported in Table \ref{adhoc2}. 
\end{itemize}

\subsection{Data collection}
Regarding the interaction evaluation, for each trial we recorded a log containing all the transitions of the FSM, tracked the actual stickers shown by the participant (ground truth) and took note of the positive/negative execution of the overall trial, with a report of occasional failures. 




\section{RESULTS}
The evaluation of the feasibility of using the humanoid robot iCub to perform a narrative cooperative task is twofold, that is quantitative and qualitative.

\subsection{Quantitative Assessment}
\subsubsection{VLM Detection} To evaluate the quality of zero-shot detection of the VLM module and its impact on the subjective engagement, we computed the percentage of success rate, represented by the level of agreement between the description given by the LLM and the image chosen by the participant. The overall agreement was $86\%$.
\subsubsection{LLM generation} To assess the contribution of the LLM in variability and in the generation of the story, we reported how many times the LLM module added additional elements to the piece invented by the participant and how many times the LLM ``fixed'' the piece added by the user. In $8\%$ of the trials, the LLM added a small contribution to the piece created by the human in their turn, whereas it never corrected or polished the human's contribution.
\subsubsection{Failure} We reported the success rate to understand how robust the system was to different failures, given the high number of different interacting modules (wrong sticker detection, microphone not detecting voice, LLM crashing, etc). Our system was pretty robust with $88\%$ success rate.

\subsection{Qualitative Assessment} 

\subsubsection{During-interaction}

\begin{itemize}
\item \textbf{Ad-hoc 1} Results indicate that the participants liked the stories invented by iCub (mean likability score $8.61 \pm 1.38$) and perceived the stories as coherent (mean coherence score $8.28 \pm 1.27$). 
\end{itemize}

\subsubsection{Post-interaction} \label{postQuest}
\begin{figure*}[t]
   \centering
          \includegraphics[width=1\textwidth]{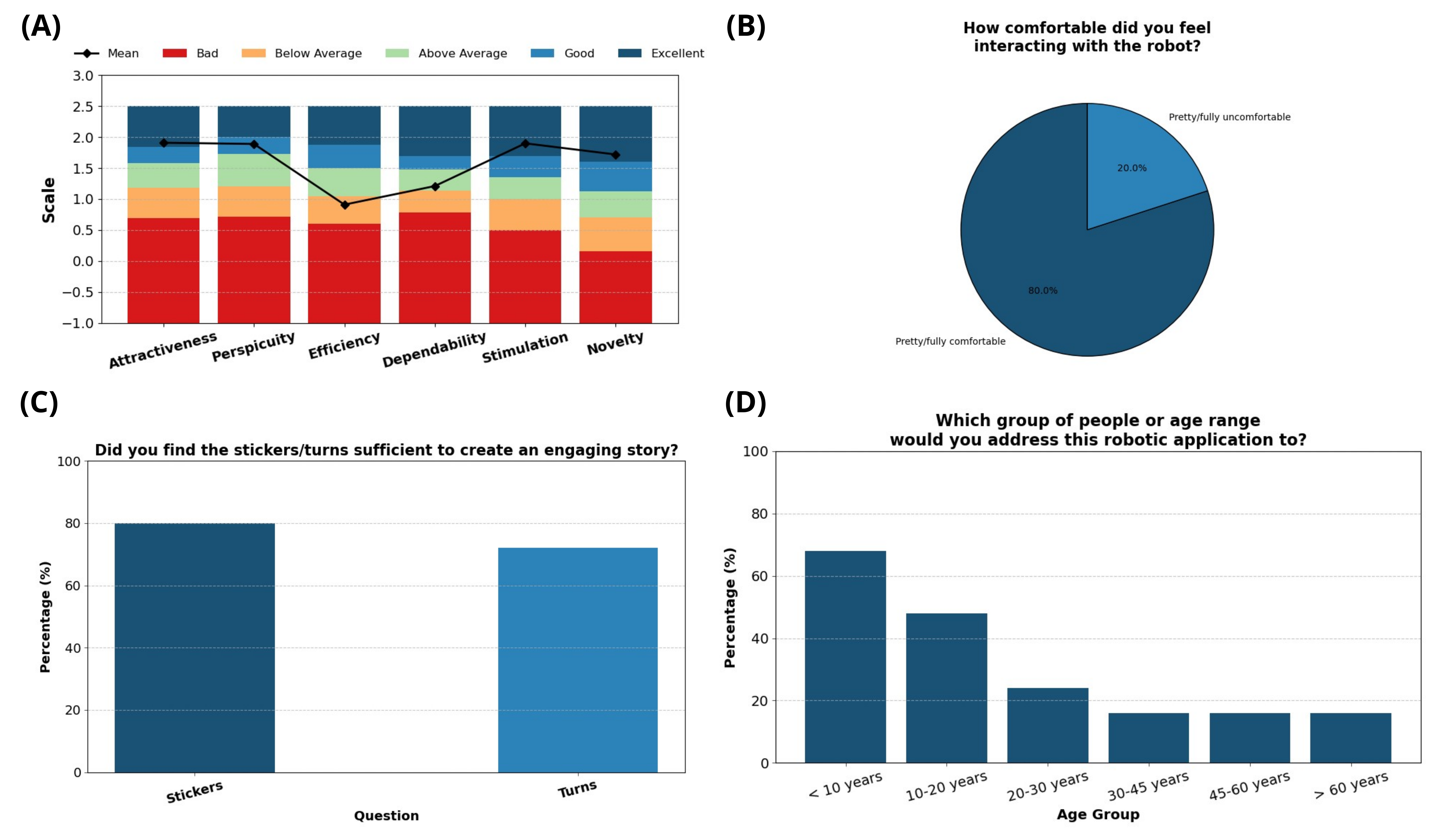}
   \caption{\textbf{Overview of questionnaires' results.} (A) The panel presents the six UEQ mean scales in comparison to the benchmark dataset associated with the questionnaire \cite{schrepp2017construction}. (B) The pie chart shows the distribution of the answers to Q1 from the Ad-hoc 2 survey. (C) The bar plot shows the participants' satisfaction level regarding stickers and turns during the storytelling (see Q2/3 from Ad-hoc 2 survey). (D) The bar plot depicts how participants would address our application based on the potential users' age- Q4 in Ad-hoc 2 questionnaire.}
   \label{fig:adhocq2}
\end{figure*}

\begin{itemize}
\item \textbf{SUS} The total score reached $76.70$, which indicates an overall good product usability according to an empirical evaluation of the questionnaire \cite{bangor2008empirical}.    
\item \textbf{UEQ} Results indicate generally positive evaluations across all six dimensions, with mean scores ranging from $0.91$ to $1.91$ (see Fig.\ref{fig:adhocq2}-\textbf{(A)}). Attractiveness received the highest rating (mean score $1.91 \pm 0.75$), suggesting a strong overall positive impression. Perspicuity (mean score $1.89 \pm 0.89$) and Stimulation (mean score $1.90\pm0.73$) were also rated highly, indicating that the system was perceived as intuitive and engaging. Novelty (mean score $1.72 \pm 0.89$) suggests users found the system  innovative. Dependability (mean score $1.21\pm0.95$) and Efficiency (mean score $0.91 \pm 0.85$) received relatively lower ratings but were still positive, implying that while the system was considered reliable and effective, there may be room for improvement in these aspects. The comparison with a reference benchmark data set \cite{schrepp2017construction} (see different bar chart colors in Fig.\ref{fig:adhocq2} for different qualitative rating categories) revealed that Attractiveness, Stimulation, and Novelty were rated as Excellent, placing them within the top $10\%$ of results. Perspicuity was classified as Good, with $75\%$ of results performing worse. Dependability was rated Above Average, with $50\%$ of results performing worse, while Efficiency was rated Below Average, indicating that $50\%$ of results were better. These findings suggest that while the system was perceived as highly engaging and visually appealing, improvements in efficiency may be beneficial. In particular, the reported perceived slowness of the interaction might have affected this aspect.


\item \textbf{Ad-hoc 2} The analysis revealed interesting features about the proposed framework (see Fig.\ref{fig:adhocq2}-\textbf{(B-D)}). Answers to Q1 highlighted that $80\%$ of the participants felt pretty/fully comfortable playing with iCub, with a very high appreciation score of the interaction (Q5: $9 \pm 1.67$ over $10$), largely willing to play with it again (Q7: $84\%$ of participants answered positively). However, we collected also feedback regarding the richness and depth of the storytelling task. The $80\%$ of the participants thought that the provided visual inputs on the cubes to start and develop a story were enough (see Q2) while the $28\%$ of players believed that having more turns in the game (in total, on average, $5 \pm 0.45$, see Q3) would have been beneficial for a more engaging storytelling. 
Furthermore, according to $68\%$ of users this application is primarily suitable for children under $10$ years old, while $46\%$ believe it is also appropriate for young people aged between $10$ and $20$ (see Q4). 
A Chi-squared goodness-of-fit test indicated that participants did not select all age groups equally ($\chi^2(5)=88.78$, $p < .001$). Pairwise two-proportion z-tests with Holm correction revealed that the below-10 group showed a significantly higher proportion of responses compared to all other age groups (all $p<=0.01$) whereas the other comparisons did not show significant results suggesting that participants were more likely to identify children under 10 years old as the intended target for this robotic application.
In the end, answers to question Q6 help us to collect feedback on how to improve the framework from the technical point of view. Many people, in fact, would have preferred a more fluent and quicker interaction with iCub (mainly due to the slow motor movement and AI-modules running on the network), with more engagement from the robot, for example with longer chatting phases, especially at the beginning as ice-breaking.
\end{itemize}

\begin{table*}[h]
\caption{\textbf{Items of the ad-hoc questionnaire 1 filled during the Interaction phase. }}\label{adhoc1}
\begin{tabular*}{\textwidth}{p{0.05\textwidth}p{0.4\textwidth}p{0.05\textwidth}p{0.4\textwidth}}
\toprule%
Item & Question & Item & Question \\
\midrule
1 & How much did you enjoy the story? Please provide a rating from 1 (``I didn’t enjoy it at all'') to 10 (``I had a lot of fun!'') & 2 & How coherent do you find the generated story? Please provide a rating from 1 (``The story made no sense at all'') to 10 (``I think the story was really fluid and coherent'')\\
\bottomrule
\end{tabular*}
\end{table*}

\begin{table*}[h]
\caption{\textbf{Items of the ad-hoc questionnaire 2 filled during the Post-Interaction phase.}}\label{adhoc2}
\begin{tabular*}{\textwidth}{p{0.05\textwidth}p{0.4\textwidth}p{0.05\textwidth}p{0.4\textwidth}}
\toprule%
Item & Question & Item & Question \\
\midrule
1 & How comfortable did you feel interacting with the robot?& 5 & How much did you enjoy the interaction with iCub?\\
2 & Did you find the stickers sufficient to create an engaging story? If not, how many would be needed? & 6 & What would you change?\\
3 & Did you find the turns sufficient to create an engaging story? If not, how many would be needed? & 7 & Would you have liked to play with iCub more? \\
4 & Which group of people or age range would you address this robotic application to? & &  \\
\bottomrule
\end{tabular*}
\end{table*}

\section{CONCLUSIONS}
Narrative architectures are often used to foster empathy
and connection towards other people. Embodied social robots equipped with LLMs conversational capabilities can facilitate emotional engagement in everyday life. In this work, we presented a setup allowing the iCub robot to perceive and understand various social cues, enabling natural, and effective HRI. Our method combines face detection, human tracking, online object detection, and generative models for verbal interaction to implement a narrative task where a human and iCub have to create a story together through cubes' exchange. This approach promises to enhance the robot’s social awareness and responsiveness, allowing for more fluid interactions, which makes it a proper tool in assistive applications, like robot-aided therapies addressed to ASD children. For this reason, before testing this solution with children, we evaluated the robotic platform with adults to validate its success rate and usability. We reported good success rate ($88\%$) regarding the technical evaluation. Participants positively rated the human-robot experience in terms of usability and attractiveness, with an overall appreciation of the interaction scored 9 over 10. Additionally, we examined the most suitable age range for this application, finding that $68\%$ of participants considered the robotic platform most acceptable and useful for children under 10. Given our goal of promoting storytelling with the iCub robot in pediatric rehabilitation, these promising results support advancing toward the integration of this novel protocol in clinical settings.

\section*{LIMITATIONS}
We acknowledge that this study raises several open questions that require further technical development to enhance the robot’s ability to effectively engage with a human partner. In particular, the results from the \textbf{Ad-hoc 2} questionnaire -reported in Section \ref{postQuest}- highlighted the need to increase the complexity of the co-created stories. This could be achieved by enriching the number of visual stimuli and increasing the interaction turns between the robot and the user, to improve the richness and depth of the story. 

Moreover, the analysis of the $12\%$ failed tests revealed specific technical issues that need to be addressed. These include the robustness of the voice transcription system, latency introduced by external services such as ChatGPT, and failures in cube detection or cube drops while interacting. The latter, for example, can be addressed by integrating visual or tactile controls so that iCub can restart the cube handover in case of failure.

Finally, it is important to note that this study was designed as a feasibility assessment of using iCub to support a (clinical) narrative protocol. As such, the experiments with adults should be considered as intermediate step. This phase was necessary to motivate and justify a subsequent study involving young patients, which requires compliance with strict regulatory protocols such as clinical evaluations.




\bibliographystyle{IEEEtran}
\bibliography{biblio} 

\end{document}